\documentclass[letterpaper, 10 pt, conference]{ieeeconf}  % Comment this line out if you need a4paper

\usepackage{cite}
\usepackage{amsmath,amssymb,amsfonts}
\usepackage{algorithmic}
\usepackage{graphicx}
\usepackage{textcomp}
\usepackage{xcolor}
\usepackage{hyperref}
\usepackage{multirow}
\usepackage{booktabs}

\IEEEoverridecommandlockouts                              % This command is only needed if 
\title{\LARGE \bf
Continuous Token-Level Spatio-Temporal Context Modeling for Visual Object Tracking
}

\author{Ding Xia$^{1}$, Meiqin Liu$^{1,2}$, Jing Zhou$^{2}$, Jian Lan$^{3}$% <-this % stops a space
\thanks{*This work is supported by the National Key R\&D Program of China under Grant 2024YFB4710600 and the National Natural Science Foundation of China under Grant U23B200653 (Corresponding author: Meiqin Liu).}% <-this % stops a space
\thanks{$^{1}$State Key Laboratory of Human-Machine Hybrid Augmented Intelligence, Xi’an Jiaotong University, Xi’an 710049, China.}%
\thanks{$^{2}$College of Electrical Engineering, Zhejiang University, Hangzhou 310027, China.}%
\thanks{$^{3}$School of Electronics and Information Engineering, Xi’an Jiaotong University, Xi’an 710049, China.}%
}

\begin{document}

\maketitle
\thispagestyle{empty}
\pagestyle{empty}

%%%%%%%%%%%%%%%%%%%%%%%%%%%%%%%%%%%%%%%%%%%%%%%%%%%%%%%%%%%%%%%%%%%%%%%%%%%%%%%%
\begin{abstract}

Spatio-temporal context has become increasingly crucial for visual tracking. However, most existing approaches extract spatio-temporal cues via discrete sampling strategies, which inherently deviate from the continuity of spatio-temporal context, thereby deteriorating tracking performance. To address this challenge, we propose TLCTrack, a novel tracking framework that models token-level spatio-temporal context through continuously updated salient tokens, enabling more accurate target representation. Specifically, TLCTrack incorporates three components: Masked Unidirectional Attention (MUA), Spatial Salient Token Collection (SSTC), and Temporal Salient Token Bank (TSTB) modules. By explicitly integrating spatio-temporal context, MUA extracts discriminative target-aware spatial features in the search region. To avoid the negative impact of background on feature learning, SSTC progressively suppresses background interference, thereby enhancing target spatial representation. Finally, TSTB captures high-quality spatio-temporal information through continuous salient token updates. Extensive experiments on five benchmarks demonstrate that our method achieves superior performance over state-of-the-art trackers. Code and models are available at \url{https://github.com/xiading123/TLCTrack}.

\end{abstract}

%%%%%%%%%%%%%%%%%%%%%%%%%%%%%%%%%%%%%%%%%%%%%%%%%%%%%%%%%%%%%%%%%%%%%%%%%%%%%%%%
\section{Introduction}
\label{sec:intro}

Visual tracking aims to estimate the trajectory of a target object throughout a video sequence given its initial state, presenting a significant challenge in computer vision \cite{SiamFC,SiamRPN++,Stark}. Traditional trackers \cite{SiamFC,SiamRPN++,Stark, TransT, AiATrack} typically employ the divide-and-conquer strategy, decomposing the tracking pipeline into two stages: feature extraction and relation modeling. Recent advancements \cite{OSTrack, SimTrack} have introduced a one-stream framework based on Vision Transformer (ViT) \cite{ViT}, which unifies two stages into a single transformer architecture. This paradigm is favored for its powerful feature learning capability.

\begin{figure}[t!]
  \centering
  \centerline{\includegraphics[width=8.5cm]{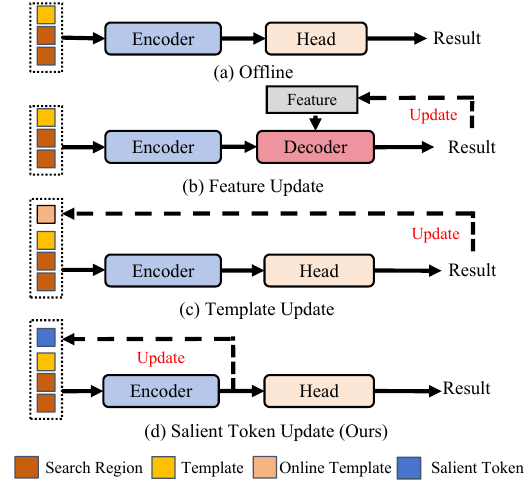}}
%  \vspace{2.0cm}
%
\caption{Comparison of different tracking pipelines. (a): Offline learning. (b)-(c): Sampling-based discrete update strategies. (d): Token-level continuous update strategy.}
\label{fig1}
\end{figure}

Despite their success, the one-stream framework still faces significant challenges in spatio-temporal context modeling. As shown in Fig. \ref{fig1}(a), most existing one-stream methods \cite{OSTrack, SimTrack} formulate visual tracking as an offline learning problem and rely on appearance modeling from template-search pairs. This approach hinders target relation modeling across search frames and fails to capture spatio-temporal context, thereby limiting tracking performance. 

To overcome this limitation, recent methods \cite{ARTrack, AQATrack, CVTTrack, SeqTrack, MixFormer, SuperSBT} have attempted to leverage spatio-temporal context. These methods can be broadly categorized into two paradigms: feature update and template update. As shown in Fig. \ref{fig1}(b), the feature update methods \cite{ARTrack, AQATrack, CVTTrack} refine tracking by updating state or motion feature over time. In contrast, as shown in Fig. \ref{fig1}(c), the template update methods \cite{MixFormer, SuperSBT, SeqTrack} capture target appearance variations and establish sparse spatio-temporal context by maintaining online templates, thereby introducing temporal cues. However, both paradigms fundamentally rely on sampling-based discrete update strategies. This coarse discrete pipeline fails to capture continuous spatio-temporal context and incurs significant computational burden. These limitations naturally raise the following question: \textbf{How to design a tracker that leverages token-level refined features to adaptively model evolving spatio-temporal context in a continuous manner?}

\begin{figure}[t!]
  \centering
  \centerline{\includegraphics[width=8.5cm]{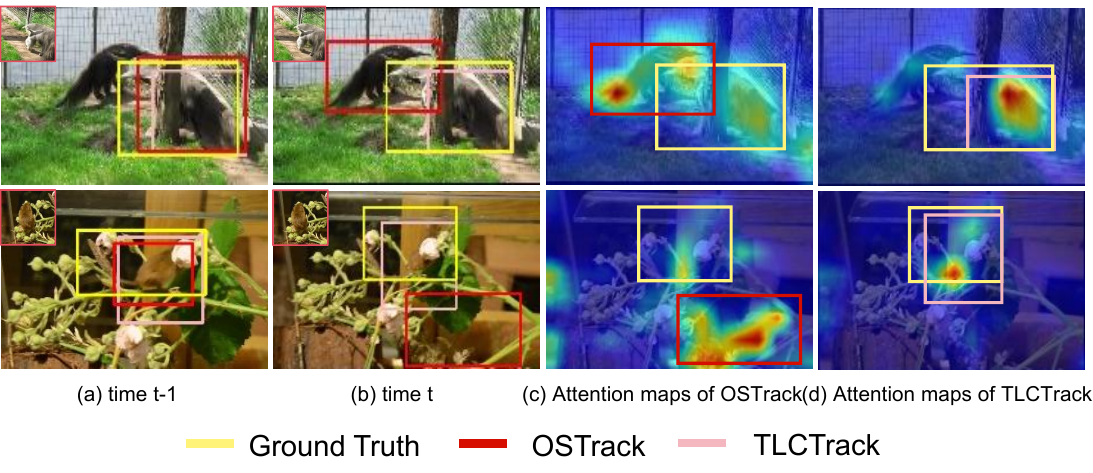}}
%  \vspace{2.0cm}
%
\caption{Comparison of the response between OSTrack and TLCTrack. TLCTrack consistently focuses on target-relevant region. (a)-(b): Search results with predicted boxes and template image (in the top-left corner). (c): Attention maps of OSTrack at time t. (d): Attention maps of TLCTrack at time t.}
\label{attnvis}
\end{figure}

To address the question, we propose \(\textbf{TLCTrack}\), a novel tracking framework that models token-level spatio-temporal context using salient tokens that store high-quality target information. As shown in Fig. \ref{fig1}(d), TLCTrack continuously collects and updates salient tokens across the video sequence, enabling progressive token-level refinement. In contrast to existing sampling-based methods, this continuous updating mechanism facilitates more effective spatio-temporal context modeling. As shown in Fig. \ref{attnvis}, the attention visualization demonstrates that TLCTrack consistently focuses on target-relevant regions, leading to better tracking performance compared to OSTrack \cite{OSTrack}. Specifically, we design three core components: the Masked Unidirectional Attention (MUA), the Spatial Salient Token Collection (SSTC), and the Temporal Salient Token Bank (TSTB) modules. To accurately aggregate target-aware spatial features in the search region, we introduce MUA to integrate spatio-temporal context of salient tokens into target features. Meanwhile, to enhance the quality of target spatial representations, SSTC is used to eliminate noisy background regions to avoid their negative impact on feature learning. Furthermore, to store high-quality spatio-temporal information, TSTB is employed to update salient tokens and filter out invalid temporal features.

The main contributions of this work are three-fold: (1) We propose TLCTrack, a tracking framework that models continuous token-level spatio-temporal context. Through the proposed TSTB and SSTC modules, TLCTrack effectively captures high-quality spatio-temporal information and constructs robust target representations. (2) We design MUA to integrate target spatio-temporal context for extracting discriminative target-aware spatial features from the search region, thereby enabling more accurate target localization and reducing drift under distractors. (3) We conduct extensive experiments and analyses on five standard benchmarks, demonstrating that TLCTrack consistently achieves state-of-the-art performance.

\begin{figure*}[t!]
  \centering
  \centerline{\includegraphics[width=16cm]{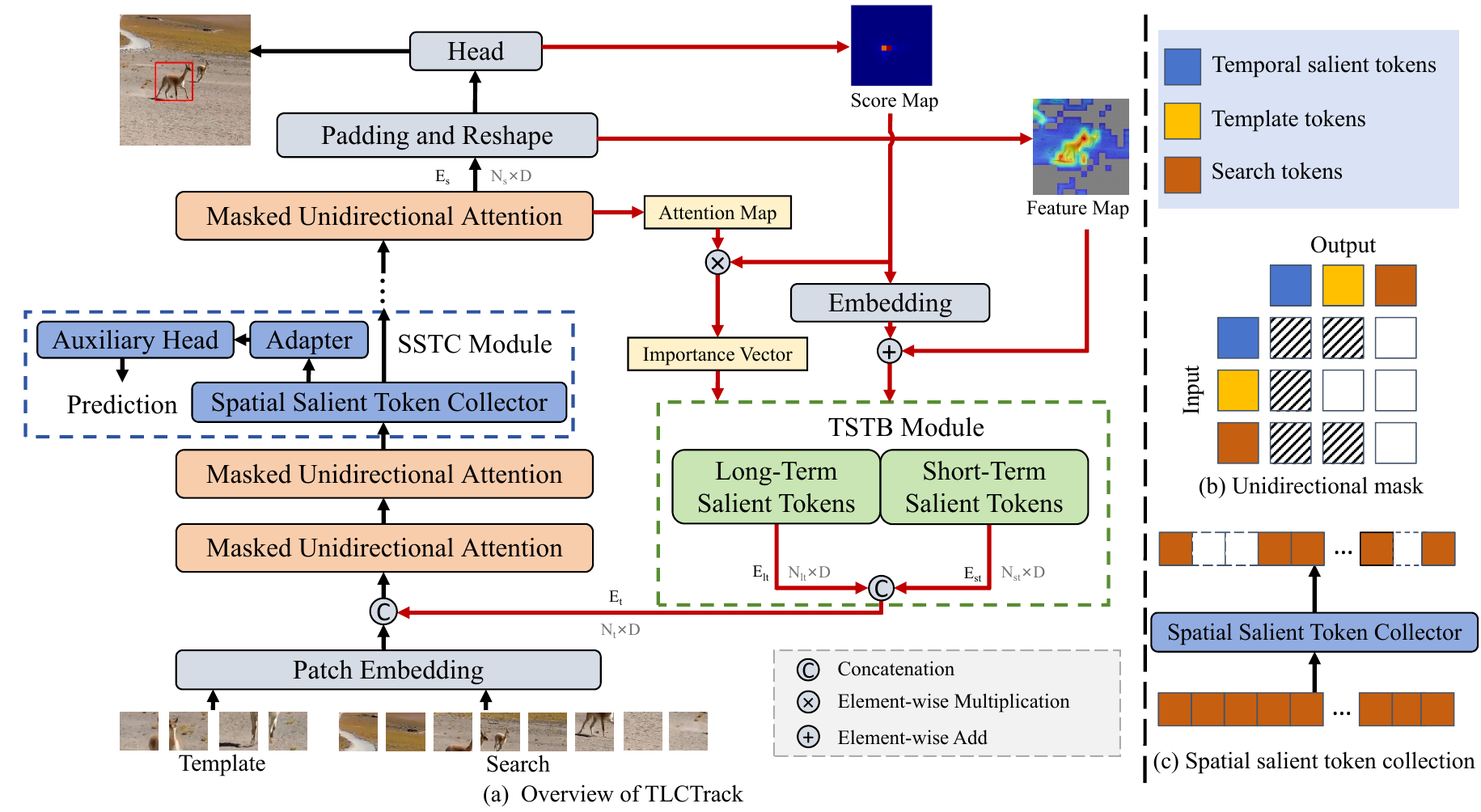}}
%  \vspace{2.0cm}
%
\caption{(a) Architecture of the proposed TLCTrack. TLCTrack consists of four parts: a backbone with MUA for visual feature extraction, SSTC for collecting spatial salient tokens, TSTB for storing temporal salient tokens, and a prediction head for making predictions. (b) Unidirectional masking among three kinds of tokens in MUA. (c) The process of collecting spatial salient tokens in SSTC.}
\label{overview}
\end{figure*}

\section{Method}
\label{sec:method}

\subsection{Overview}
\label{ssec:Overview}

The architecture of TLCTrack is shown in Fig. \ref{overview}(a). It consists of four parts: a backbone with MUA for feature extraction, SSTC for collecting spatial salient tokens, TSTB for storing temporal salient tokens, and a prediction head for making predictions. To capture continuous token-level spatio-temporal context across frames, we introduce two types of salient tokens: temporal salient tokens \(\mathbf{E}_{\text{t}}\) (the concatenation of long-term salient tokens \(\mathbf{E}_\text{lt}\) and short-term salient tokens \(\mathbf{E}_\text{st}\)) and spatial salient tokens \(\mathbf{E}_\text{s}\).

TLCTrack takes a template and a search region as input. First, the template and search region are converted into template tokens \(\mathbf{E}_z \in \mathbb{R}^{N_z \times D}\) and search tokens \(\mathbf{E}_x \in \mathbb{R}^{N_x \times D}\) through patch embedding. Then, these tokens are concatenated with \(\mathbf{E}_{\text{t}} \in \mathbb{R}^{N_{\text{t}} \times D}\) from TSTB along the spatial dimension and fed into the backbone. Acting as a plug-in module, SSTC operates within these blocks to collect \(\mathbf{E}_\text{s} \in \mathbb{R}^{N_\text{s} \times D}\). Here, \(N_z\), \(N_x\), \(N_{\text{t}}\), and \(N_\text{s}\) indicate the token number of \(\mathbf{E}_z\), \(\mathbf{E}_x\), \(\mathbf{E}_{\text{t}}\), and \(\mathbf{E}_\text{s}\), while \(D\) indicates the dimension of each token. Subsequently, TSTB collects and updates \(\mathbf{E}_{\text{t}}\) to capture token-level temporal features. Finally, the backbone outputs search features, which are forwarded to the prediction head to compute the tracking results.

\subsection{Masked Unidirectional Attention Module}
\label{ssec:MUA}

To accurately aggregate target-aware spatial features in the search region under the guidance of temporal information, we propose MUA to regulate information flow through unidirectional masking. As shown in Fig. \ref{overview}(b), we employ multi-head attention (MHA) with unidirectional mask to define relationship modeling rules across three token categories: \(\mathbf{E}_z\), \(\mathbf{E}_x\), \(\mathbf{E}_{\text{t}}\). Let \(\boldsymbol{q}\), \(\boldsymbol{k}\), and \(\boldsymbol{v}\) represent the queries, keys and values input into the \(l\)-th multi-head attention layer. The relation modeling rules for these three token categories are formulated below. 

\(\mathbf{E}_z\) can only aggregate information via self-attention:
\begin{equation}
\label{eq1}
\boldsymbol{q}_z = \boldsymbol{k}_z = \boldsymbol{v}_z = \mathbf{E}_{z}^{l},
\end{equation}
\begin{equation}
\label{eq2}
\mathbf{E'}_{z}^{l} = \mathbf{E}_{z}^{l} + \mathrm{MHA}(\boldsymbol{q}_z, \boldsymbol{k}_z, \boldsymbol{v}_z),
\end{equation}
\(\mathbf{E}_x\) can aggregate information from all token categories:
\begin{equation}
\label{eq3}
\mathbf{\boldsymbol{q}}_x = \mathbf{E}_{x}^{l}, 
\quad
\mathbf{\boldsymbol{k}}_x = \mathbf{\boldsymbol{v}}_x = [\mathbf{E}_{\text{t}}^{l}; \mathbf{E}_{z}^{l}; \mathbf{E}_{x}^{l}],
\end{equation}
\begin{equation}
\label{eq4}
\mathbf{E'}_{x}^{l} = \mathbf{E}_{x}^{l} + \mathrm{MHA}(\boldsymbol{q}_x, \boldsymbol{k}_x, \boldsymbol{v}_x),
\end{equation}
For \(\mathbf{E}_{\text{t}}\), these tokens are constrained to remain invariant in MUA as follows:
\begin{equation}
\label{eq5}
\mathbf{E'}_{\text{t}}^{l} = \mathbf{E}_{\text{t}}^{l}.
\end{equation}

Information from \(\mathbf{E}_{\text{t}}\) and \(\mathbf{E}_z\) is unidirectionally integrated into \(\mathbf{E}_x\) via the attention operations in (\ref{eq3}) and (\ref{eq4}), while MUA keeps \(\mathbf{E}_{\text{t}}\) unchanged to maintain temporal independence among them from different frames as shown in (\ref{eq5}).

The MUA mechanism enables unidirectional information propagation from \(\mathbf{E}_{\text{t}}\) to \(\mathbf{E}_x\) while prohibiting interactions within \(\mathbf{E}_{\text{t}}\), thereby effectively maintaining the integrity and independence of the temporal features. This design extracts discriminative target-aware spatial features under the guidance of temporal information, thereby enabling more accurate target localization.

\subsection{Spatial Salient Token Collection Module}
\label{ssec:SSTC}

Prior work \cite{GRM} utilizes masked relation modeling to improve foreground token quality by weakening background noise contamination of the target features, but this approach cannot fully block noise interference. Inspired by \cite{Cropr}, we propose SSTC that progressively collects spatial salient tokens \(\mathbf{E}_\text{s}\) to enhance target spatial representations, as shown in Fig. \ref{overview}(c).

First, we define a score vector \(\boldsymbol{a}\) to quantify the importance of candidate search tokens \(\mathbf{E}_x\), which is then used to collect \(\mathbf{E}_\text{s}\). Specifically, we introduce learnable queries \(\mathbf{Q} \in \mathbb{R}^{N_x \times D}\) to construct \(\boldsymbol{a}\). To simplify notation, we denote candidate search tokens $\mathbf{E}_x$ as \(\mathbf{X} \in \mathbb{R}^{N \times D}\), where $N \le N_x$ due to previous collection. \(\boldsymbol{a} \in \mathbb{R}^{1 \times N}\) is defined as follows:
\begin{equation}
\label{eq6}
\boldsymbol{a} = \sum_{i=1}^{N_x} (\mathbf{Q}\,\mathbf{X}^{\top})_{i,:}.
\end{equation}

Then, based on \(\boldsymbol{a}\), \(\mathbf{E}_\text{s}\) can be collected as:
\begin{equation}
\label{eq7}
\mathcal{I}_\text{s} = \operatorname{arg-TopK}(\boldsymbol{a}, N_\text{s}),
\quad
\mathbf{E}_\text{s} = \mathbf{X}\big[\mathcal{I}_\text{s}\big],
\end{equation}
where \(N_\text{s}\) is a hyperparameter, with the token keeping ratio defined as \(\rho = {N_\text{s}}/{N}\) and \(\mathcal{I}_\text{s}\) denotes the indices of high-score tokens.

To optimize the learnable queries \(\mathbf{Q}\), we introduce an adapter and an auxiliary head. The adapter integrates information from \(\mathbf{Q}\) and \(\mathbf{X}\), defined as follows:
\begin{equation}
\label{eq8}
\mathbf{X'} = \mathrm{Softmax}\left( \frac{\mathbf{Q}\,\mathbf{X}^{\top}}{\sqrt{D}} \right)\mathbf{X},
\end{equation}
\begin{equation}
\label{eq9}
\mathbf{X''} = \mathbf{X'} + \mathrm{MLP}\left(\mathbf{X'}\right).
\end{equation}

The auxiliary head takes \(\mathbf{X''}\) as input to make predictions, which in turn provides gradients to train the adapter and \(\mathbf{Q}\). We compute the auxiliary loss and add it to the overall loss with a discount factor \(\lambda_{\text{aux}}\). To decouple token collection from feature extraction, we apply a stop-gradient operation to detach \(\mathbf{X}\) before (\ref{eq8}).

We utilize SSTC to effectively filter out background noise and improve the quality of target spatial representations, thereby enhancing the robustness of tracking.

\begin{table*}[!t]
\caption{Comparison with state-of-the-arts on four benchmarks: GOT10k, TrackingNet, TNL2K and NFS. * denotes for trackers only trained on GOT10k. Best in bold, second best underlined.}
\centering
\begin{tabular}{l|ccc|ccc|cc|c}
\toprule
% \multicolumn{1}{c|}{Method} & \multicolumn{3}{c}{GOT10k} \\
\multirow{2}{*}{Method} & \multicolumn{3}{c|}{GOT10k*} & \multicolumn{3}{c|}{TrackingNet} & \multicolumn{2}{c|}{TNL2K} & \multicolumn{1}{c}{NFS} \\
\cmidrule(lr){2-4}\cmidrule(lr){5-7}\cmidrule(lr){8-9}\cmidrule(lr){10-10}
& \text{AO(\%)} & $\text{SR}_{0.5}\text{(\%)}$ & $\text{SR}_{0.75}\text{(\%)}$ & \text{AUC(\%)} & $\text{P}_{Norm}\text{(\%)}$ & $\text{P(\%)}$ & \text{AUC(\%)} & $\text{P(\%)}$ & \text{AUC(\%)} \\
\midrule
OSTrack$_{384}$ \cite{OSTrack} & 73.7 & 83.2 & 70.8 & 83.9 & 88.5 & 83.2 & 55.9 & - & 66.5 \\
SeqTrack$_{256}$ \cite{SeqTrack} & 74.7 & 84.7 & 71.8 & 83.3 & 88.3 & 82.2 & 54.9 & - & 67.6 \\
SeqTrack$_{384}$ \cite{SeqTrack} & 74.5 & 84.3 & 71.4 & 83.9 & 88.8 & 83.6 & 56.4 & - & 66.7 \\
ARTrack$_{256}$ \cite{ARTrack} & 73.5 & 82.2 & 70.9 & 84.2 & 88.7 & 83.5 & 57.5 & - & 64.3 \\
ARTrack$_{384}$ \cite{ARTrack} & 75.5 & 84.3 & 74.3 & \textbf{85.1} & 89.1 & \textbf{84.8} & \underline{59.8} & - & 66.8 \\
DropTrack$_{384}$ \cite{DropMAE} & 75.9 & 86.8 & 72.0 & 84.1 & 88.9 & - & 56.9 & 57.9 & - \\
MixFormer$_{384}$ \cite{MixFormer} & 75.7 & 85.3 & \underline{75.1} & 83.9 & 88.9 & 83.1 & - & - & - \\
HIPTrack$_{384}$ \cite{HIPTrack} & \underline{77.4} & \underline{88.0} & 74.5 & 84.5 & 89.1 & 83.8 & - & - & \underline{68.1} \\
AQATrack$_{256}$ \cite{AQATrack} & 73.8 & 83.2 & 72.1 & 83.8 & 88.6 & 83.1 & 57.8 & 59.4 & - \\
AQATrack$_{384}$ \cite{AQATrack} & 76.0 & 85.2 & 74.9 & \underline{84.8} & \underline{89.3} & \underline{84.3} & 59.3 & \underline{62.3} & - \\
ADAT$_{384}$ \cite{ADAT} & - & - & - & 84.2 & 88.9 & 83.7 & 58.4 & 60.1 & - \\
\midrule
\textbf{TLCTrack}$_{\textbf{256}}$ & 75.4 & 86.3 & 74.3 & 83.7 & 88.6 & 82.7 & 57.0 & 58.4 & 67.3  \\
\textbf{TLCTrack}$_{\textbf{384}}$ & \textbf{77.5} & \textbf{88.4} & \textbf{76.1} & \underline{84.8} & \textbf{89.6} & \underline{84.3} & \textbf{60.1} & \textbf{62.7} & \textbf{70.7} \\
\bottomrule
\end{tabular}
\label{Comparison-with-SOTA}
\end{table*}

\subsection{Temporal Salient Token Bank Module}
\label{ssec:TSTB}

Prior works \cite{ARTrack, SuperSBT} leverage features extracted from recent frames to address target appearance variations over time. However, their discrete sampling approaches fail to capture continuous spatio-temporal context. To address this limitation, TSTB collects short-term salient tokens \(\mathbf{E}_\text{st} \in \mathbb{R}^{N_\text{st} \times D}\) to capture appearance variations from the recent frame, while continuously updating long-term salient tokens \(\mathbf{E}_\text{lt} \in \mathbb{R}^{N_\text{lt} \times D}\) to selectively retain high-quality temporal features. Initially, \(\mathbf{E}_\text{lt}\) and \(\mathbf{E}_\text{st}\) are updated from the template tokens \(\mathbf{E}_z\), thus both \(N_\text{lt}\) and \(N_\text{st}\) are equal to \(N_z\). Then, \(\mathbf{E}_\text{lt}\) and \(\mathbf{E}_\text{st}\) are concatenated along the spatial dimension to form the temporal salient tokens \(\mathbf{E}_{\text{t}}\) as follows:
\begin{equation}
\label{eq10}
\mathbf{E}_{\text{t}} = [\mathbf{E}_\text{lt}; \mathbf{E}_\text{st}].
\end{equation}

 To continuously collect and update \(\mathbf{E}_\text{lt}\) and \(\mathbf{E}_\text{st}\), we define an importance criterion for salient tokens. Specifically, let \(\mathbf{Q}_\text{s}\), \(\mathbf{K}_{\text{s}}\) and \(\mathbf{K}_\text{t}\) denote the query and keys from \(\mathbf{E}_{\text{s}}\) and \(\mathbf{E}_\text{t}\) in the final layer.

\(\mathbf{E}_\text{lt}\) is updated from the current \(\mathbf{E}_{\text{t}}\) based on its importance score vector \(\boldsymbol{m}_{\text{t}} \in \mathbb{R}^{1 \times N_{\text{t}}}\). The update procedure of \(\mathbf{E}_\text{lt}\) is defined as:
\begin{equation}
\label{eq11}
\boldsymbol{m}_{\text{t}} = \sum_{i=1}^{N_\text{s}}(\mathrm{Softmax}\left( \frac{\mathbf{Q}_\text{s}\,\mathbf{K}_{\text{t}}^{\top}}{\sqrt{D}} \right) \times \mathbf{S})_{i,:},
\end{equation}
\begin{equation}
\label{eq12}
\mathcal{I}_\text{lt} = \operatorname{arg-TopK}(\boldsymbol{m}_{\text{t}}, N_\text{lt}),
\quad
\mathbf{E'}_\text{lt} = \mathbf{E}_{\text{t}}\big[\mathcal{I}_\text{lt}\big],
\end{equation}
where \(\mathbf{S}\) is the score map from the prediction head and \(\mathcal{I}_\text{lt}\) denotes the indices of high-score tokens.

\(\mathbf{E}_\text{st}\) is obtained from the feature map \(\mathbf{F'}\), which aggregates \(\mathbf{E}_\text{s}\) with the prediction result. First, we define \(\mathbf{F}\) as the backbone output feature map, constructed by padding and reshaping \(\mathbf{E}_\text{s}\), and \(\mathbf{S}_{\text{bin}}\) as the binary score map converted from \(\mathbf{S}\). \(\mathbf{F'}\) is formulated as:
\begin{equation}
\label{eq13}
\mathbf{F'} = \mathrm{Embedding}(\mathbf{S}_{\text{bin}}) + \mathbf{F},
\end{equation}

Then, \(\mathbf{E}_\text{st}\) is updated based on importance score vector \(\boldsymbol{m}_\text{s} \in \mathbb{R}^{1 \times N_{\text{s}}}\). The update procedure of \(\mathbf{E}_\text{st}\) is defined as:
\begin{equation}
\label{eq14}
\boldsymbol{m}_\text{s} = \sum_{i=1}^{N_\text{s}}(\mathrm{Softmax}\left( \frac{\mathbf{Q}_\text{s}\,\mathbf{K}_\text{s}^{\top}}{\sqrt{D}} \right) \times \mathbf{S})_{i,:},
\end{equation}
\begin{equation}
\label{eq15}
\mathcal{I}_\text{st} = \operatorname{arg-TopK}(\boldsymbol{m}_\text{s}, N_\text{st}),
\quad
\mathbf{E'}_\text{st} = \mathbf{F'}\big[\mathcal{I}_\text{st}\big],
\end{equation}
where \(\mathcal{I}_\text{st}\) denotes the indices of high-score tokens.

Subsequently, \(\mathbf{E'}_\text{lt}\) and \(\mathbf{E'}_\text{st}\) are concatenated to form \(\mathbf{E'}_{\text{t}}\) as follows:
\begin{equation}
\label{eq16}
\mathbf{E'}_{\text{t}} = [\mathbf{E'}_\text{lt}; \mathbf{E'}_\text{st}],
\end{equation}
where \(\mathbf{E}_\text{lt}\) progressively preserves high-quality temporal information across frames while eliminating invalid features, and \(\mathbf{E}_\text{st}\) selectively extracts the latest target appearance information. By adaptively integrating both \(\mathbf{E}_\text{lt}\) and \(\mathbf{E}_\text{st}\), TSTB continuously updates \(\mathbf{E}_{\text{t}}\) to maintain spatio-temporal context, enabling temporal information to guide future prediction and improving robustness against distractors.

\subsection{Head and Loss}
\label{ssec:Head and Loss}

Following \cite{OSTrack}, both the auxiliary head and the prediction head consist of three convolutional branches, which output the classification score map, local offset, and bounding box size. We employ the focal loss as classification loss \(L_{cls}\) \cite{FocalLoss}, and the \(L_1\) loss and \(GIoU\) loss \cite{GIOU} as regression loss. For each head, the loss function is defined as:
\begin{equation}
\label{eq17}
L = L_{cls} + \lambda_{L1} L_{1} + \lambda_{iou} L_{GIoU},
\end{equation}
where \(\lambda_{L1}\) = 5 and \(\lambda_{iou}\) = 2. The loss of the auxiliary head is incorporated into the overall loss with a discount factor \(\lambda_{\text{aux}}\) of 0.5.

\section{Experiments}
\label{sec:experiments}

\subsection{Implementation Details}
\label{ssec:Implementation}

\(\textbf{Model}\). We use ViT-Base \cite{ViT} model pre-trained with MAE \cite{MAE} as the backbone. Following \cite{OSTrack}, the proposed SSTC is added at blocks 4, 7, and 10, with the keeping ratio \(\rho\) of 0.8. To demonstrate the scalability of TLCTrack, we design two variants with the following configurations:

\textbullet\ \(\text{TLCTrack}_{256}\). Template size: [128$\times$128]; Search region size: [256$\times$256];

\textbullet\ \(\text{TLCTrack}_{384}\). Template size: [192$\times$192]; Search region size: [384$\times$384].

\begin{table}[!t]
	\caption{Comparison of model parameters, FLOPs, inference speed, and memory overhead on an RTX 3090 GPU.}
	\label{Comparison-speed}
	\centering
	{
	\begin{tabular}[t]{c|ccccc}  
		\toprule
		Method  & Params $\downarrow$ & FLOPs $\downarrow$ & Speed $\uparrow$ & Memory $\downarrow$ \\
		\midrule
		SeqTrack$_{256}$ &  89M  & 66G & 52\,\textit{fps} & 1653M \\
		SeqTrack$_{384}$ &  89M  & 148G & 22\,\textit{fps} & 3855M \\
		ARTrack$_{256}$ &  173M  & 37G & 62\,\textit{fps} & 1287M \\
		ARTrack$_{384}$ &  173M  & 83G & 43\,\textit{fps} & 1343M \\
		\midrule
		\textbf{TLCTrack}$_{\textbf{256}}$ &  92M  & 28G & 97\,\textit{fps} & 951M \\
		\textbf{TLCTrack}$_{\textbf{384}}$ &  92M  & 58G & 67\,\textit{fps} & 1122M \\ 
		\bottomrule
	\end{tabular} }
\end{table}

\(\textbf{Training}\). Following traditional protocols, the training splits of TrackingNet \cite{TrackingNet}, LaSOT \cite{LaSOT}, COCO \cite{COCO}, and GOT10k \cite{GOT10k} (remove 1,000 videos as \cite{Stark}) are used for training. For evaluation on GOT10k test set, we exclusively use its training set, adhering to the official requirements \cite{GOT10k}.
To simulate spatio-temporal context transmission, we construct each batch with three template frames and four search frames. First, two templates pass through the backbone to initialize \(\mathbf{E}_{\text{t}}\). Next, the third template is sequentially paired with each search frame; each pair is processed by the model, with \(\mathbf{E}_{\text{t}}\) updated iteratively. The losses from all four search frames are then computed for backpropagation. We employ AdamW \cite{AdamW} with a weight decay of \(10^{-4}\) to optimize the network parameters, with the initial learning rates of \(2\times10^{-5}\) for the backbone and \(2\times10^{-4}\) for the rest. The model is trained for 300 epochs with 60k image pairs, and the learning rates are reduced by a factor of 10 after 240 epochs. For GOT10k, the model is trained for 100 epochs, and the learning rates are decayed at the 80th epoch. Training is carried out on two RTX 4090 GPUs, with a total batch size of 64.

\(\textbf{Inference}\). In the initial stage of tracking, both \(\mathbf{E}_\text{lt}\) and \(\mathbf{E}_\text{st}\) in \(\mathbf{E}_{\text{t}}\) are updated using the tokens from the template. Then, the template and the current search frame are input into the trained model sequentially. Further, we perform a comparative evaluation of model parameters, FLOPs, inference speed and memory overhead on an RTX 3090 GPU, as shown in Table \ref{Comparison-speed}.

\begin{figure}[t!]
  \centering
  \centerline{\includegraphics[width=9.0cm]{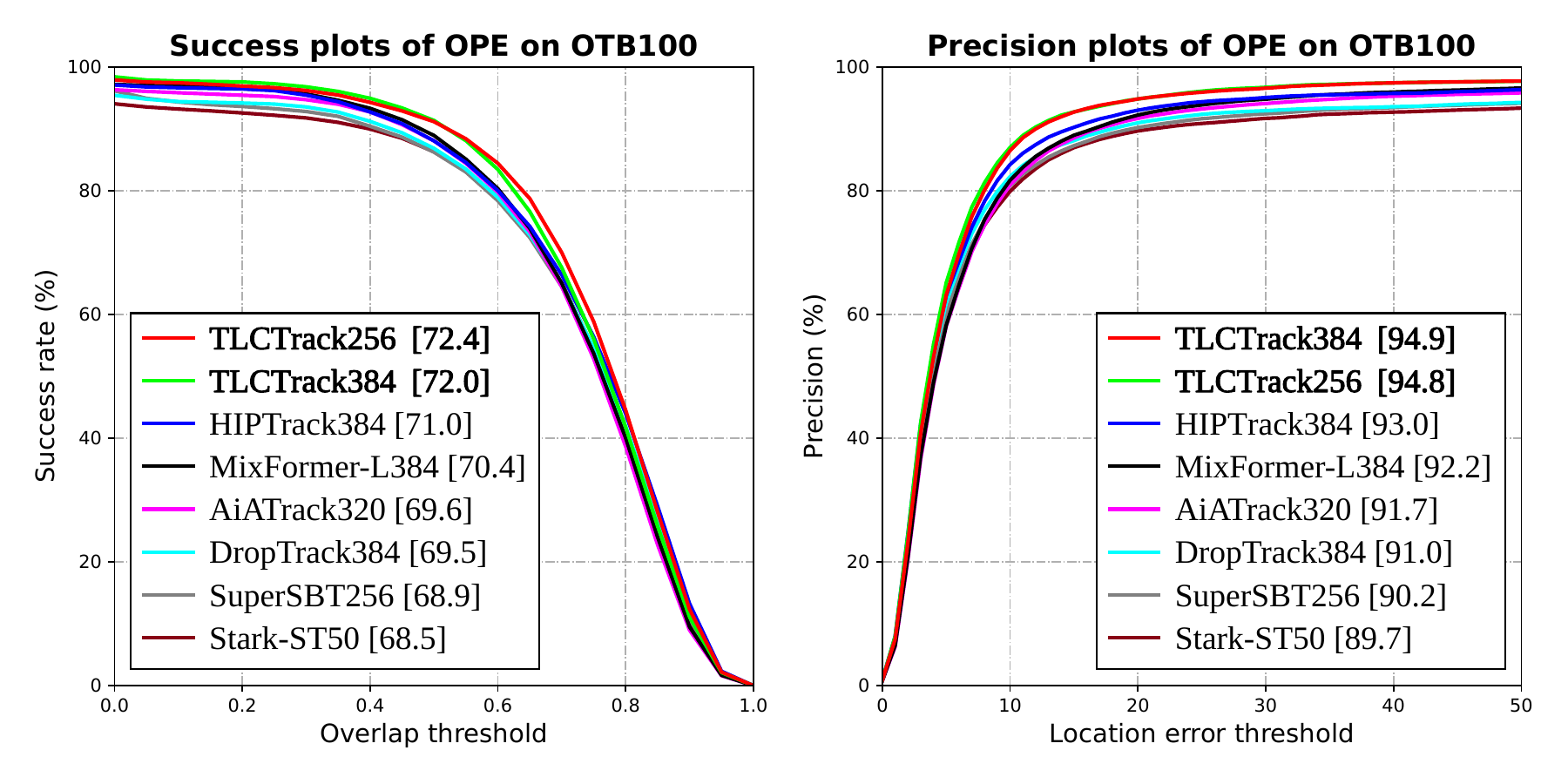}}
%  \vspace{2.0cm}
%
\caption{State-of-the-art comparison on the OTB100 dataset. TLCTrack outperforms previous SOTA trackers. Best viewed with zooming in.}
\label{Results-on-OTB}
\end{figure}

\subsection{Results and Comparisons}
\label{ssec:Performance Comparison}

To demonstrate the effectiveness of our method, TLCTrack is compared with the previous SOTA trackers on five benchmarks, including GOT10k \cite{GOT10k}, TrackingNet \cite{TrackingNet}, TNL2K \cite{TNL2K}, NFS \cite{NFS}, OTB100 \cite{OTB}.

\(\textbf{GOT10k}\). GOT10k is a comprehensive dataset comprising over 10,000 video sequences, with 180 sequences designated for testing. As reported in Table \ref{Comparison-with-SOTA}, \(\text{TLCTrack}_{384}\) achieves state-of-the-art performance of 77.5\% AO, outperforming previous state-of-the-art trackers. These outcomes demonstrate the strong performance of TLCTrack and validate the effectiveness in modeling spatio-temporal context.

\(\textbf{TrackingNet}\). TrackingNet comprises 511 testing sequences. Table \ref{Comparison-with-SOTA} shows that TLCTrack delivers competitive performance against most prior trackers. Notably, Table \ref{Comparison-speed} reports that \(\text{TLCTrack}_{384}\) runs at 67$\,\textit{fps}$, substantially faster than \(\text{ARTrack}_{384}\) (43$\,\textit{fps}$). These results validate that our method achieves a superior trade-off between tracking accuracy and computational efficiency.

\(\textbf{TNL2K}\). TNL2K serves as a comprehensive benchmark containing 700 difficult video sequences. As shown in Table \ref{Comparison-with-SOTA}, \(\text{TLCTrack}_{384}\) achieves a new state-of-the-art score of 60.1\% in \text{AUC} and 62.7\% in \text{P}, surpassing prior best results.

\(\textbf{NFS}\). NFS comprises 100 video sequences, totaling 380,000 video frames. The results in Table \ref{Comparison-with-SOTA} demonstrate that TLCTrack outperforms previous state-of-the-art approaches, achieving 70.7\% in \text{AUC}.

\(\textbf{OTB100}\). OTB100 contains 100 tracking sequences covering eleven common challenges. The results in Fig. \ref{Results-on-OTB} show that our approach surpasses state-of-the-art methods on OTB100, demonstrating the strong robustness of TLCTrack.

\subsection{Qualitative Results}
\label{ssec:Qualitative Results}
\begin{figure}[t!]
  \centering
  \centerline{\includegraphics[width=9.0cm]{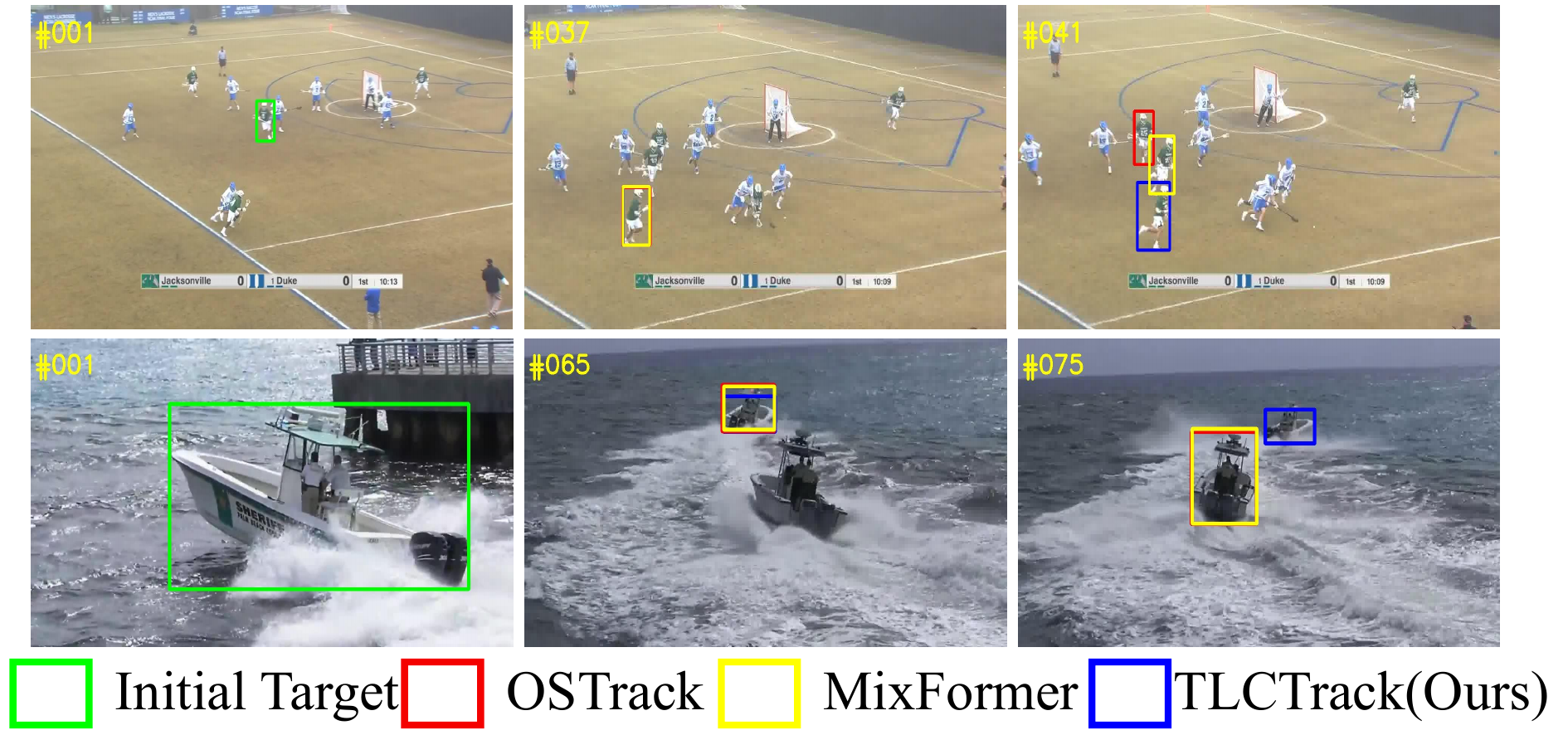}}
%  \vspace{2.0cm}
%
\caption{Qualitative evaluation with other leading trackers (000078 and 000139 from GOT10k test split). TLCTrack consistently maintains stable target localization. Best viewed with zooming in.}
\label{Qualitativevis}
\end{figure}

\textbf{Visualization}. Qualitative comparisons are provided in Fig. \ref{Qualitativevis}, where our approach is evaluated against two leading trackers, OSTrack \cite{OSTrack} and MixFormer \cite{MixFormer}. As shown, when facing challenging conditions such as highly cluttered backgrounds and target appearance variations, OSTrack and MixFormer suffer from tracking failures. In contrast, TLCTrack consistently maintains accurate and stable localization. These outcomes indicate that leveraging spatio-temporal context via salient tokens effectively improves tracking robustness.

\begin{table*}[!t]
\caption{Three ablation studies on the GOT10k benchmark.}
\centering
\begin{tabular}{ccccc|cccc|cccc}
\toprule
\multicolumn{5}{c|}{(a) Study on key components} & \multicolumn{4}{c|}{(b) Study on keeping ratio \(\rho\)} & \multicolumn{4}{c}{(c) Study on the loss weight \(\lambda_{\text{aux}}\)} \\
\midrule
\# & Method & \text{AO(\%)} & $\text{SR}_{0.5}\text{(\%)}$ & $\text{SR}_{0.75}\text{(\%)}$ & \(\rho\) & \text{AO(\%)} & $\text{SR}_{0.5}\text{(\%)}$ & $\text{SR}_{0.75}\text{(\%)}$ & \(\lambda_{\text{aux}}\) & \text{AO(\%)} & $\text{SR}_{0.5}\text{(\%)}$ & $\text{SR}_{0.75}\text{(\%)}$ \\
\midrule
1 & baseline & 72.0 & 81.6 & 69.5 & 0.7 & 74.0 & 83.6 & 72.8 & 0.1 & 74.3 & 84.5 & 73.2 \\ 
2 & +TSTB & 73.3 & 82.7 & 71.6 & 0.8 & \textbf{75.4} & \textbf{86.3} & \textbf{74.3} & 0.3 & 74.9 & 85.7 & 73.8 \\
3 & +MUA & 74.6 & 84.8 & 73.2 & 0.9 & 74.9 & 85.3 & 73.7 & 0.5 & \textbf{75.4} & \textbf{86.3} & \textbf{74.3} \\
4 & +SSTC & \textbf{75.4} & \textbf{86.3} & \textbf{74.3} & 1.0 & 74.6 & 84.8 & 73.2 & 0.7 & 74.6 & 85.3 & 73.7 \\
\bottomrule
\end{tabular}
\label{Ablation-studies}
\end{table*}

\subsection{Ablation Study and Analysis}
\label{ssec:Ablation Studies}

\textbf{Component analysis}. We evaluate the cumulative contribution of these modules in \(\text{TLCTrack}_{256}\) on GOT10k, as detailed in Table \ref{Ablation-studies}(a). Our baseline is a plain ViT implementation (\#1). First, TSTB establishes token-level temporal features, leading to a 1.3\% AO gain (\#1 and \#2). Building upon this foundation, MUA more effectively leverages spatio-temporal information through masking attention, outperforming naive cross-attention (\#2 and \#3). Finally, SSTC further refines target spatial representation, achieving 75.4\% AO (\#4). Overall, this progressive design, from establishing spatio-temporal context to optimizing its integration and finally refining spatial features, demonstrates strong synergy and leads to substantial performance improvements.

\textbf{Study on Keeping Ratio \(\rho\)}. We conduct ablation studies to evaluate the impact of the keeping ratio \(\rho\) in SSTC on model performance, as detailed in Table \ref{Ablation-studies}(b). Note that \(\rho=1\) means SSTC is not adopted. Table \ref{Ablation-studies}(b) shows that the model achieves peak performance when the keeping ratio is set at 0.8. A lower ratio may overlook important target-aware spatial features, while a higher ratio fails to adequately suppress background noise.

\textbf{Study on the loss weight \(\lambda_{\text{aux}}\)}. We conduct ablation studies on the loss weight \(\lambda_{\text{aux}}\) of the auxiliary head, as detailed in Table \ref{Ablation-studies}(c). As shown, optimal performance is achieved at \(\lambda_{\text{aux}}=0.5\). Appropriate \(\lambda_{\text{aux}}\) enables SSTC to learn more discriminative target spatial representations, thereby collecting \(\mathbf{E}_\text{s}\) accurately. In contrast, improper \(\lambda_{\text{aux}}\) degrades the generalization capabilities of SSTC.

\section{Conclusion}
\label{sec:conclusion}

In this paper, we propose TLCTrack, a novel tracking framework that models continuous spatio-temporal context by continuously updating salient tokens. At its core, we design three modules, including MUA, SSTC, and TSTB, to effectively leverage spatio-temporal information through continuous salient token updates across the video sequence. Specifically, MUA unidirectionally integrates temporal information into search region, SSTC enhances the quality of target spatial representation, and TSTB continuously captures high-quality spatio-temporal context. Experimental results on five benchmarks demonstrate that TLCTrack achieves superior performance over previous state-of-the-art trackers, exhibiting remarkable robustness and capability.

\bibliographystyle{IEEEbib}
\bibliography{smc2026references}

\end{document}